\documentclass[runningheads]{llncs}

\usepackage{eccv}

\usepackage{eccvabbrv}

\usepackage{graphicx}
\usepackage{booktabs}

\usepackage[accsupp]{axessibility}  

\usepackage{hyperref}

\usepackage{orcidlink}
\usepackage{multirow}
\usepackage{rotating}
\usepackage{makecell}
\usepackage{caption}

\begin{document}

\title{SOS: Segment Object System for Open-World Instance Segmentation With Object Priors}

\titlerunning{SOS: Segment Object System}

\author{Christian Wilms\inst{1}\orcidlink{0009-0003-2490-7029} \and
Tim Rolff\inst{1,2}\orcidlink{0000-0001-9038-3196} \and
Maris Hillemann\inst{1} \and
Robert Johanson\inst{1} \and
Simone Frintrop\inst{1}\orcidlink{0000-0002-9475-3593} }


\authorrunning{C.~Wilms et al.}

\institute{Computer Vision Group, University of Hamburg, Germany
 \and
Human-Computer Interaction Group, University of Hamburg, Germany\\
\email{\{firstname.lastname\}@uni-hamburg.de}}

\maketitle

\begin{abstract} 

We propose an approach for Open-World Instance Segmentation (OWIS), a task that aims to segment arbitrary unknown objects in images by generalizing from a limited set of annotated object classes during training. Our Segment Object System~(SOS) explicitly addresses the generalization ability and the low precision of state-of-the-art systems, which often generate background detections. To this end, we generate high-quality pseudo annotations based on the foundation model SAM~\cite{kirillov2023segment}. We thoroughly study various object priors to generate prompts for SAM, explicitly focusing the foundation model on objects. The strongest object priors were obtained by self-attention maps from self-supervised Vision Transformers, which we utilize for prompting SAM. Finally, the post-processed segments from SAM are used as pseudo annotations to train a standard instance segmentation system. Our approach shows strong generalization capabilities on COCO, LVIS, and ADE20k datasets and improves on the precision by up to 81.6\% compared to the state-of-the-art. Source code is available at:~\url{https://github.com/chwilms/SOS}
  \keywords{Open-world Instance Segmentation \and Object Localization \and Prompting}
\end{abstract}

\section{Introduction}
\label{sec:intro}

\begin{figure}[]
\centering
\includegraphics[width=\linewidth]{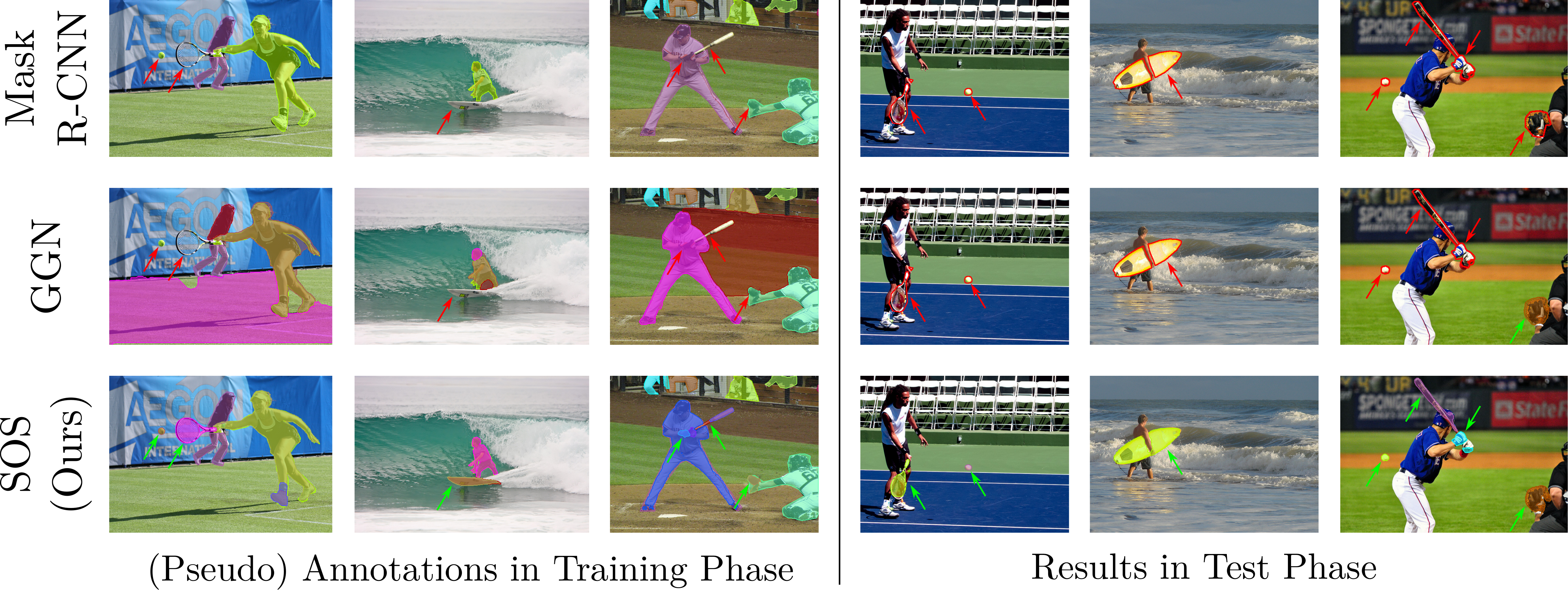}
\caption{Comparison of (pseudo) annotations (left) used by Mask R-CNN~\cite{he2017mask}, GGN~\cite{wang2022open}, and  our SOS, when only original annotations of VOC object classes are given. While Mask R-CNN only uses original annotations without object classes such as tennis racket or surfboard~(red arrows), SOS generates pseudo annotations covering those classes~(green arrows). GGN generates noisy pseudo annotations including  background areas. As a result, only SOS constantly detects these objects not annotated in training~(green vs. red arrows on the right). Filled masks denote annotations~(left) or detected objects~(right), while red frames indicate missed objects.}
\label{fig:intro}
\end{figure}

Open-World Instance Segmentation (OWIS) is the task of segmenting all object instances in an image by learning from a limited set of known object classes~\cite{kim2022learning,wang2022open,wang2021unidentified}. In contrast to instance segmentation, OWIS is not limited by the closed-world assumption, which assumes that all object classes are known in advance.
Since OWIS methods aim to detect not only learned but also unknown object classes, they return class-agnostic detections. This is of particular interest in real-world scenarios with previously unknown object classes~(e.g., \cite{xie2021unseen}) and challenges the systems' generalization capabilities. An example of this is shown in the first row of~\cref{fig:intro}, where annotations for classes such as surfboard or tennis racket do not exist during training, but the objects should be detected during testing.

Previous OWIS methods mostly generate pseudo annotations for unannotated objects~\cite{wang2022open,saito2022learning,kalluri2023open,xue2022single} or replace the foreground-background classification of possible objects in instance segmentation systems with a learning target focusing on the localization quality (e.g., intersection over union)~\cite{kim2022learning,wu2023exploring,wang2023openinst}. 
While OWIS methods generalize better to unseen object classes than standard instance segmentation systems, they still exhibit low precision~\cite{wu2023exploring,kim2022learning,wang2022open,saito2022learning,kalluri2023open}.
One reason for this is the use of noisy pseudo annotations covering background areas as visible in the left part of the second row in~\cref{fig:intro}.

Recently, foundation models~\cite{bommasani2021opportunities} emerged as a promising technique~\cite{brown2020language,devlin2018bert,jia2021scaling,radford2021learning,li2023blip}. They are trained on large datasets with a surrogate task and applied to other tasks in a zero- or few-shot manner using prompting.  The first foundation model for image segmentation, the Segment Anything Model (SAM), was proposed by~\cite{kirillov2023segment}. Trained on a large automatically annotated dataset, SAM generates segments for object or stuff regions based on point, box, mask, or text prompts without further training. Despite not only segmenting objects but also stuff regions, given well-designed prompts, SAM is able to generate high-quality object segments. Note that vanilla SAM does not address object segmentation tasks like OWIS itself due to the aforementioned lacking focus on objects.

In this paper, we propose the \emph{Segment Object System}~(SOS), a novel OWIS method utilizing high-quality pseudo annotations to improve the generalization capability of a standard instance segmentation system~(see lower row in~\cref{fig:intro}). For generating pseudo annotations, we apply the foundation model SAM with prompts derived from an object prior to roughly localize objects of arbitrary classes and to limit segmentations of stuff regions by SAM. To ensure the high quality of our pseudo annotations based on SAM, improving the overall precision, we thoroughly study various hand-crafted and learned object priors for an object-focused application of SAM. This is relevant beyond the scope of OWIS. Using our study findings, SOS utilizes self-attention maps from self-supervised Vision Transformers (ViTs)~\cite{dosovitskiy2020image} as object prior for prompting SAM. Finally, SOS filters low-quality pseudo annotations, combines the pseudo annotations with the original annotations of the known classes, and trains a standard instance segmentation system with these mixed annotations. Our extensive evaluation shows the strong generalization abilities of SOS, considerably outperforming all previous state-of-the-art systems across COCO~\cite{lin2014microsoft}, LVIS~\cite{gupta2019lvis}, and ADE20k~\cite{zhou2019semantic} datasets. Most notably, the results improve by up to 81.6\% in terms of precision over the state-of-the-art due to the high-quality pseudo annotations. We also show that SOS is better suited for OWIS than directly applying SAM. 

Overall, our contributions are threefold:
\begin{itemize}
    \item We propose SOS, a novel OWIS method based on a learned object prior, prompt-based pseudo annotations, and an arbitrary instance segmentation system.
    \item We thoroughly study various object priors for focusing SAM on objects, leading to high-quality object segments for  pseudo annotations and beyond.
    \item Our extensive evaluation shows that the high-quality pseudo annotations in SOS lead to high precision and strong overall results, clearly outperforming other methods.
\end{itemize}

\section{Related Work}

\subsection{Open-world Instance Segmentation}
To address OWIS, literature mainly offers two streams. First, systems replace the hard classification in instance segmentation systems with a localization score. This avoids classifying unseen objects as background and improves generalization ability~\cite{kim2022learning}. In OLN~\cite{kim2022learning}, centerness and Intersection over Union (IoU) are learned as localization scores in a Mask R-CNN system~\cite{he2017mask}. SWORD~\cite{wu2023exploring} and OpenInst~\cite{wang2023openinst} follow the same idea for query-based  systems.

The second line of works~\cite{wang2022open,saito2022learning,kalluri2023open,xue2022single} addresses OWIS by augmenting the annotations from known classes with pseudo annotations aiming to cover objects of unknown classes. To generate pseudo annotations, GGN~\cite{wang2022open}~learns pixel affinities from known classes to generate segments, which are grouped. Similarly, UDOS~\cite{kalluri2023open}~groups object proposals of object parts to create pseudo annotations, while LDET~\cite{saito2022learning} applies copy-paste-augmentation. Recently, a new stream emerged~\cite{zhu2023segprompt,yang2024class} explicitly utilizing class labels from known classes.

We follow the second stream and create pseudo annotations within SOS, since it offers the most flexibility w.r.t.~the base instance segmentation system. Different from existing approaches that generate noisy pseudo annotations (see GGN in~\cref{fig:intro}), we investigate and develop object priors to effectively use recent foundation models to create high-quality object-focused pseudo annotations, leading to high recall and precision.


\subsection{Class-agnostic Object Localization}
Localizing objects in images is related to several computer vision tasks. For instance, in traditional object proposal generation~\cite{alexe2010object}, cues like saliency~\cite{alexe2010object,garcia2015saliency} or contour information~\cite{alexe2010object,rahtu2011learning,dollar2013structured,ma2017object} were used to localize class-agnostic object candidates for object detection. Since the advent of deep learning, the models are learned from data~\cite{o2015learning,girshick2015fast,pinheiro2016learning,hu2017fastmask,wilms2019attentionmask}. 

Focusing on arbitrary salient objects, known as salient object detection, several approaches use the contrast of hand-crafted features such as color~\cite{cheng2014global,perazzi2012saliency,liu2010learning,klein2011center} or edge information~\cite{klein2011center}, as well as CNN-based approaches~\cite{hou2017deeply,liu2018picanet,zhao2019egnet}. In a different task, several authors investigated the information stored in learned classifiers or self-supervised feature extractors based on CNNs or ViTs to locate objects. Notable avenues include Class Activation Maps~(CAMs)~\cite{zhou2016learning} that highlight discriminative object or image parts in CNN-based classifiers and the self-attention maps in the self-supervised DINO feature extractor~\cite{caron2021emerging} based on ViTs, which contain information on the scene layout indicating object locations.

In our study on object priors to focus SAM on objects, we investigate several of these object localization cues. In contrast to these approaches, we follow a different goal, focusing prompt-based segmentations on objects for OWIS.

\subsection{Applications of SAM}
SAM~\cite{kirillov2023segment} is used in several segmentation tasks~\cite{baugh2023zero,shi2024training,mazurowski2023segment}. Similar to us, \cite{chen2023segment} and \cite{yang2024foundation} use SAM to generate pseudo annotation masks for weakly-supervised semantic segmentation based on image-level supervision. Similarly, \cite{he2023weakly}~generate pseudo annotation masks for concealed object segmentation based on scribble annotations. Finally, \cite{wei2023semantic} generate pseudo annotation masks for weakly-supervised instance segmentation in a multiple instance learning framework.

Different from the weakly-supervised approaches utilizing SAM, we do not rely on supervision. Hence, a key novelty of our work lies in investigating object priors to focus SAM on segmenting arbitrary objects in the absence of supervision for all object classes. Moreover, we are the first to apply SAM-based pseudo annotations in OWIS.

\section{Segment Object System for OWIS}
\label{sec:method}
\begin{figure}[t]
    \centering
    \includegraphics[width=\linewidth]{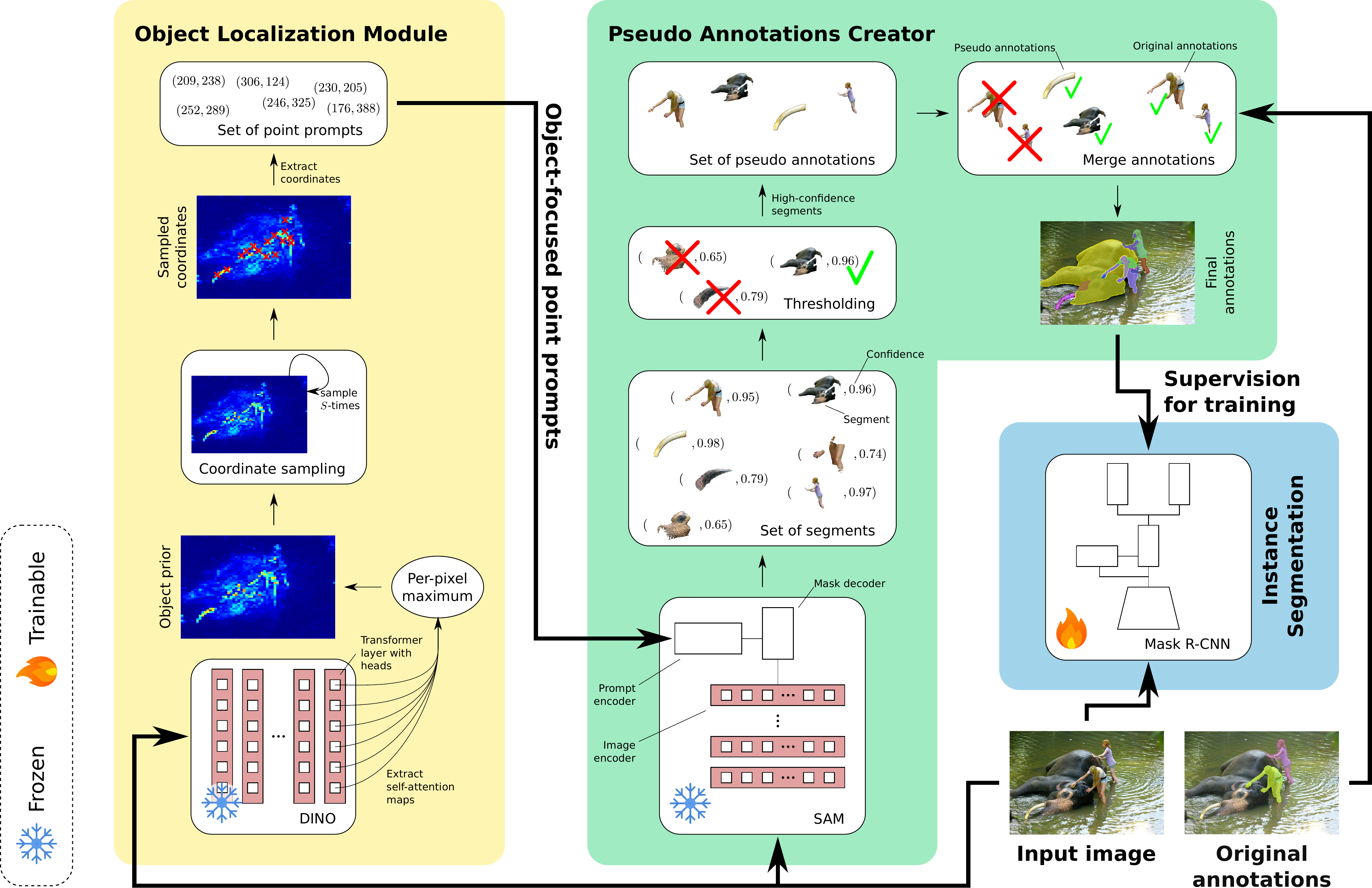}
    \caption{Overview of our Segment Object System (SOS) for OWIS consisting of three blocks. First, the input image is processed in our Object Localization Module~(OLM, yellow area) to create object-focused point prompts roughly localizing objects. Second, our Pseudo Annotations Creator~(PAC, green area) generates segments based on  the previously generated prompts using SAM~\cite{kirillov2023segment}, and further processes them, leading to a final set of merged original and pseudo annotations. Finally, the merged annotations are used to train an instance segmentation system~(blue area). }
    \label{fig:sysFig}
\end{figure}

To address OWIS, we propose our \emph{Segment Object System}~(SOS) that generates high-quality pseudo annotations to train a standard instance segmentation system. As visible from~\cref{fig:sysFig}, SOS consists of three main blocks. The first block, the Object Localization Module~(OLM), described in~\cref{sec:method_olm}~(yellow area in~\cref{fig:sysFig}), aims to roughly localize objects by sampling locations from an object prior that reflects the object probability per image coordinate. The output of OLM is a set of likely object locations, that is used in the Pseudo Annotation Creator~(PAC), described in~\cref{sec:method_pac}~(green area in~\cref{fig:sysFig}), to prompt the pre-trained SAM~\cite{kirillov2023segment}, which we briefly revisit in~\cref{sec:method_sam}. From the segments generated from likely object locations, the PAC generates pseudo annotations by filtering low-quality segments and removing near-duplicates. Finally, the pseudo annotations are merged with the original annotations, and a standard instance segmentation system is trained with the merged annotations~(see~\cref{sec:method_is}, blue area in~\cref{fig:sysFig}). During testing, only this instance segmentation system is used.

\subsection{Revisit Segment Anything Model}
\label{sec:method_sam}
Since SAM~\cite{kirillov2023segment} is integral to SOS, we will briefly review the model. The idea of SAM is to utilize prompts such as point coordinates to generate segments of image parts like objects, object parts, or stuff regions localized by these prompts. Hence, SAM does not exclusively segment objects unless explicitly instructed by object-focused prompts, as proposed in this paper. SAM is based on a ViT~\cite{dosovitskiy2020image} image encoder creating an image embedding, while the embeddings for point prompts are generated based on positional encodings. Both embeddings are processed in a mask decoder utilizing a transformer decoder block and a mask prediction head to create the final segment including a confidence score. Note that SAM is class-agnostic and does not classify the segments.

For training SAM, \cite{kirillov2023segment}~propose the new large-scale SA-1B dataset that is annotated with a data engine using SAM itself. In the pre-stage, SAM is trained with public segmentation datasets to aid the first stage of the data engine, where annotators use SAM as an interactive segmentation tool to create initial segmentations. Subsequently, SAM is retrained with the newly annotated data, fully avoiding data leakage w.r.t.~public segmentation datasets. In subsequent stages, manual annotations were provided for segments missed by SAM. At last, annotations were created fully automatically by SAM. See~\cite{kirillov2023segment} for more details.


\subsection{Object Localization Module}
\label{sec:method_olm}
The first block of SOS~(yellow area in~\cref{fig:sysFig}) is the Object Localization Module~(OLM).  OLM roughly localizes all objects in the input image, and creates object-focused point prompts. As the first step,  OLM generates an object prior from the input image. The object prior is a probability mass function constrained to the image plane, although other formulations are possible~(see~\cref{sec:objPrior}), and indicates the probability of an image coordinate being part of any object. Hence, the object prior is class-agnostic and not limited to known object classes from the training set in OWIS. In our study in~\cref{sec:objPrior}, we analyze various object priors. For final SOS, we  utilize the six self-attention maps from the self-attention heads of a ViT's final layer, pre-trained in the self-supervised DINO framework~\cite{caron2021emerging} on ImageNet~\cite{russakovsky2015imagenet}. See~\cref{sec:objPrior} for justifications. OLM aggregates the six self-attention maps by elementwise max yielding one map highlighting all relevant scene elements. Subsequently, OLM converts the map to an object prior, by rescaling such that the minimum equals 0 and the sum over the object prior is 1.

Next, OLM randomly samples $S$~(here: $S=50$) image coordinates from the object prior, given the per-coordinate values of the object prior as probabilities. To diversify the samples across multiple objects, we prune parts of the object prior around the sampled coordinate. Specifically, given a coordinate~$(x_s,y_s)$, we set the object prior to zero for all coordinates 
$\{(x,y)\ |\ |x-x_s| \leq N\land |y-y_s|\leq N\}$,
with $N=20$. To transform the object prior back to a probability mass function, we again apply the rescaling as described above. Subsequently, OLM iteratively applies this sampling until a set of $S$ coordinates are extracted, representing object-focused point prompts. This set is the output of the OLM and will be used for prompting in the Pseudo Annotation Creator.


\subsection{Pseudo Annotations Creator}
\label{sec:method_pac}
The Pseudo Annotations Creator~(PAC, green area in~\cref{fig:sysFig}) is the second block of SOS. It generates pseudo annotations utilizing SAM, given the object-focused point prompts provided by OLM. These object-focused point prompts lifts SAM from a system that segments anything to a system that segments objects. As the first step, PAC prompts a pre-trained SAM with the $S$ point prompts from OLM. To handle ambiguous point prompts that may indicate multiple objects or object parts, we follow~\cite{kirillov2023segment} and allow SAM to generate three segments per prompt. Each generated segment is a potential pseudo annotation. The resulting set of class-agnostic segments is noisy, with segments covering objects and background, or covering the same image area. Therefore, PAC first utilizes SAM's confidence score per segment and removes segments with a confidence below~$\tau_{\text{conf}}$. Second, PAC removes near-duplicates by non-maximum suppression with an IoU threshold~$\tau_\text{NMS}$. We set $\tau_{\text{conf}}=0.9$ and $\tau_\text{NMS}=0.95$ for SOS. 

Given this set of pseudo annotations, PAC merges the set with the original annotations representing known classes of the training dataset to combine the knowledge of the human annotators and the knowledge of our object-focused SAM. Since we only want to keep pseudo annotations for objects not covered by the original annotations, we suppress pseudo annotations that have a high IoU~($\tau_\text{NMS}$) with at least one original annotation. Moreover, we limit the number of pseudo annotations to $P$ in order to balance original and pseudo annotations. In our final SOS, we set $P=10$. However, across the COCO training dataset, only 7.8 pseudo annotations are added on average. The other pseudo annotations were suppressed in previous steps. Overall, the output of  PAC is a mixed set of annotations, containing all original annotations augmented with high-quality pseudo annotations covering objects of unknown classes.


\subsection{Instance Segmentation}
\label{sec:method_is}
In the last block of SOS, we take the merged annotations generated by PAC and train a standard instance segmentation system. While this system can be arbitrary and is generally replaceable, we choose Mask R-CNN~\cite{he2017mask} with a ResNet-50+FPN backbone~\cite{lin2017feature} trained in a class-agnostic setting, following recent OWIS methods~\cite{kim2022learning,wang2022open,saito2022learning,kalluri2023open}. Overall, this leads to an OWIS method based on high-quality pseudo annotations.


\section{Object Priors for SOS}
\label{sec:objPrior}
As described in~\cref{sec:method_olm}, SOS uses an object prior for rough object localization and subsequent point prompt generation for PAC. This section thoroughly investigates the performance of various object priors derived from previous class-agnostic object localization works in SOS. For details on the object priors beyond the subsequent descriptions, we refer to our supplementary.


\subsection{Object Priors}

\paragraph{\textbf{Baselines:}}
We introduce three baseline object priors that ignore the image content. First, \textit{Grid} uses SAM with a regular $64 \times 64$ grid of points, following~\cite{kirillov2023segment}. This directly leads to prompts without the sampling described in~\cref{sec:method_olm}. Note that the \textit{Grid} baseline will lead to segments for both object and stuff regions. Second, we use the spatial distribution of object centroids from the known classes across the training dataset as object prior~(\textit{Dist}). We assume that this generalizes well since it ignores class-specific object features. Finally, we utilize the centroids of the training set's objects from unknown classes as optimal prompts~(\textit{GT}), representing an upper bound.


\paragraph{\textbf{Superpixels:}}
Inspired by~\cite{wilms2017edge,gao2017saliency}, we utilize the centroids of superpixels generated with an adaptive superpixel segmentation method as object prior~(\textit{Spx}). The intuition is that large uniform areas not containing objects should be covered by few superpixels and vice versa.  Hence, the density of superpixels is a surrogate for the density of objects. We use the well-known FH superpixels~\cite{felzenszwalb2004efficient} since they adapt their density to the image content following our assumption. Note that every superpixel centroid is a point prompt.


\paragraph{\textbf{Contour Density:}}
Since objects are defined by their outer contours~\cite{zitnick2014edge}, we use contour density as an object prior~(\textit{Contour}). Similar to~\cite{wilms2017edge}, we assume that in areas of high contour density, several objects are located and vice versa. As a surrogate for the contour density, we use edge density based on~\cite{dollar2013structured}.


\paragraph{\textbf{Saliency:}}
Saliency is used in object proposal generation to localize objects~\cite{alexe2010object,garcia2015saliency} as they stick out from their surrounding. Here, we evaluate two saliency methods as object priors. First, we apply the traditional approach VOCUS2~\cite{frintrop2015traditional} based on color contrast and used by~\cite{garcia2015saliency}~(\textit{VOCUS2}). Second, we use pre-trained DeepGaze~IIE~\cite{linardos2021deepgaze} saliency maps learned from eye fixation data~(\textit{DeepGaze}).


\paragraph{\textbf{Class Activation Maps:}}
CAMs~\cite{zhou2016learning} indicate discriminative image regions for CNN-based classifiers. Since they only need image-level supervision, they are frequently applied in weakly-supervised tasks to locate objects~\cite{ahn2019weakly,liu2020leveraging,kim2022beyond}. We use the CAM of the predicted class in a ResNet-50 classifier~\cite{he2016deep} pre-trained on ImageNet~\cite{russakovsky2015imagenet} as object prior~(\textit{CAM}). 


\paragraph{\textbf{Self-attention:}}
Recently, \cite{caron2021emerging}~have shown that self-attention maps learned inside self-supervised ViTs encode the scene layout, including object locations. Therefore, we explore the self-attention maps from the final layer of a ViT-S backbone trained in the self-supervised DINO framework~\cite{caron2021emerging} as an object prior~(\textit{DINO}). 




\paragraph{\textbf{Learned Object Locations:}}
Finally, we learn object locations based on the known classes of a dataset using a U-Net~\cite{ronneberger2015u} in a binary segmentation task~(object vs. no object) as object prior~(\textit{U-Net}). This assumes that, on the level of individual pixels, the model generalizes from known to unknown classes~\cite{wang2022open,xue2022single}. 


\subsection{Study Setup for Object Priors}
\label{sec:objPrior_eval_setup}
To assess each object prior, we train SOS with the respective prior on the COCO training set using only annotations from the 20 PASCAL VOC~\cite{everingham2015pascal} classes. We evaluate on COCO's validation set with annotations from the remaining 60 classes in COCO, following standard COCO $\text{(VOC)}\rightarrow\text{COCO}$ (non-VOC) cross-category evaluation in OWIS~\cite{saito2022learning,wang2022open,kim2022learning}. Note that we use the default class-agnostic Mask R-CNN in SOS but reduce the training schedule to only a quarter of steps for faster training. We evaluate the results of SOS given each object priors and report Average Recall (AR) for 100 detections and Average Precision (AP) as previous OWIS research~\cite{saito2022learning,wang2022open}. We also report F$_1$ score, the harmonic mean between AR and AP, yielding a single number for comparison. 


\subsection{Results of Object Priors in SOS}
\label{sec:objPrior_eval}
Table~\ref{tab:study_quan_res} shows the results of our object prior study. First, all priors outperform the baselines \textit{Grid} and \textit{Dist} in terms of F$_1$. Both baselines exhibit a lower AP compared to all other object priors, reflecting the missing focus on objects. Hence, simply applying SAM (\textit{Grid}) is not suitable to segment only objects, as several stuff regions are segmented as well. Second, most priors exhibit a similar performance in terms of all measures. For instance, learning-based \textit{CAM} and \textit{DeepGaze} do not outperform simple priors~(\textit{Spx}, \textit{Contour}, \textit{VOCUS2}). Only the priors \textit{U-Net} and \textit{DINO} perform substantially better, with \textit{DINO} producing the best result. Comparing \textit{DINO} to \textit{GT} reveals that \textit{DINO} is able to recall almost the same amount of objects, but exhibits a lower precision, as expected.

\begin{figure}[t]
    \centering
    \begin{tabular}{cccc}
\multirow{2}{*}[3.5em]{\includegraphics[height=3cm]{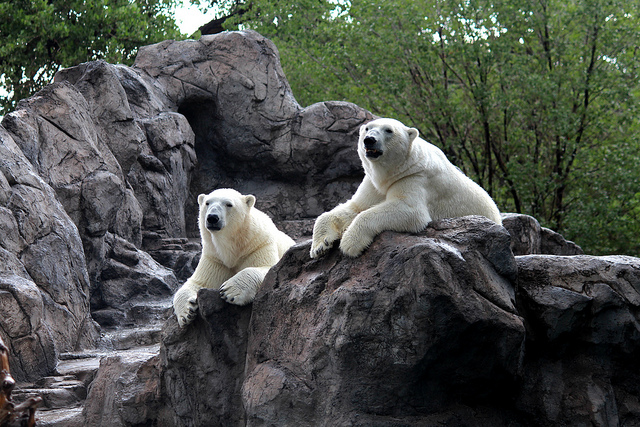}} & \includegraphics[height=1.5cm]{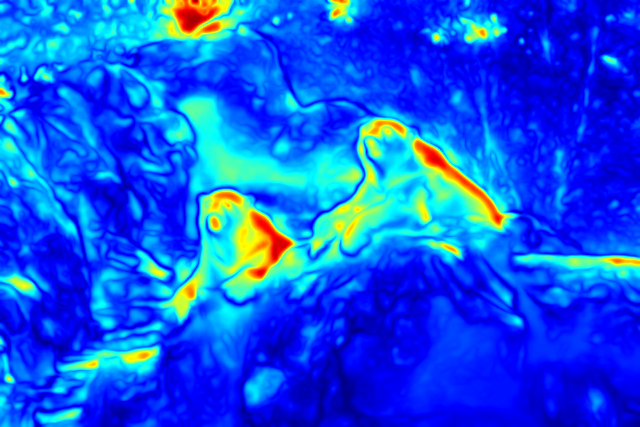} & \includegraphics[height=1.5cm]{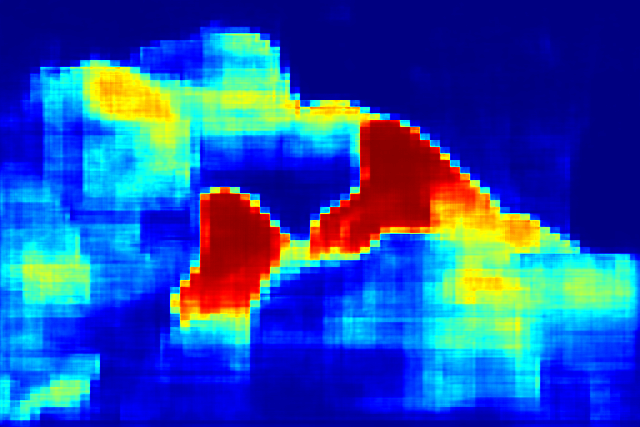} & \includegraphics[height=1.5cm]{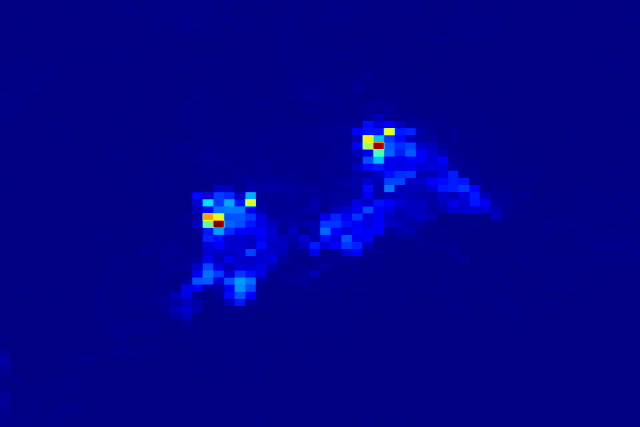} \\
 & \includegraphics[height=1.5cm]{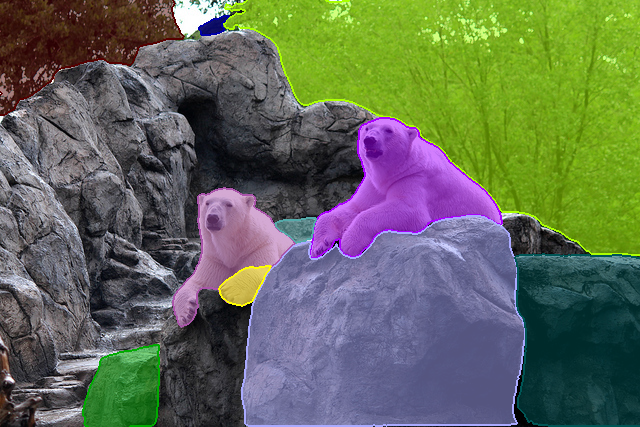} & \includegraphics[height=1.5cm]{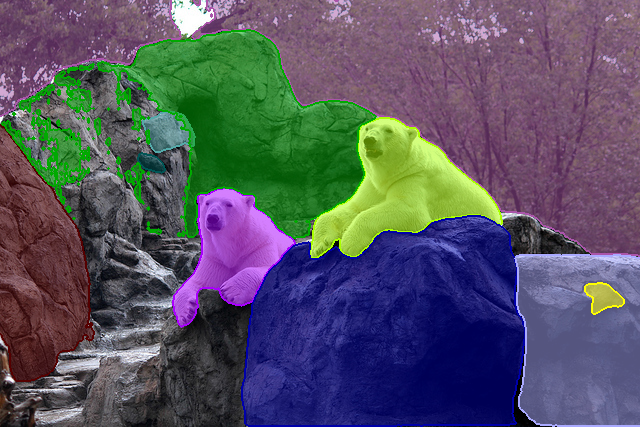} & \includegraphics[height=1.5cm]{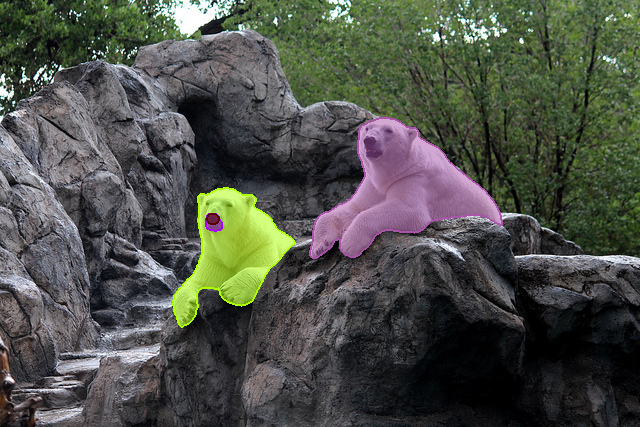} \\
Input image & \textit{VOCUS2} & \textit{U-Net} & \textit{DINO}
\end{tabular}
    \caption{Object priors \textit{VOCUS2}, \textit{U-Net}, and \textit{DINO} with resulting pseudo annotations.}
    \label{fig:qual_priors}
\end{figure}

To illustrate the results, \cref{fig:qual_priors} depicts object priors for top-performing \textit{VOCUS2}, \textit{U-Net}, and \textit{DINO}, as heatmaps~(upper row) with the resulting pseudo annotations generated in SOS~(lower row) for an example image. The heatmaps show that \textit{DINO} is very focused on the discriminative object parts, here the polar bears' faces, while \textit{U-Net} highlights, as learned, the entire objects. Moreover, \textit{U-Net} also highlights several background areas. Similarly, \textit{VOCUS2} highlights background areas and the polar bears since it is genrally focused on high-contrast areas. Overall, only \textit{DINO} exclusively generates pseudo annotations for the polar bears and some sub-parts, while \textit{U-Net} and \textit{VOCUS2} produce noisy annotations including background patches, similar to previous approaches~\cite{wang2022open}.

The results show that a well-designed object prior like \textit{DINO} substantially outperforms the baseline \textit{Grid}~(+7.5 in F$_1$) in SOS. Therefore, we use \textit{DINO} as the object prior in SOS. More generally, the study also reveals important insights on how to prompt SAM for class-agnostic, object-focused results for arbitrary tasks. For more study results, we refer to our supplementary.


\begin{table}[t]
\centering
\caption{Results of our SOS with various object priors in the cross-category COCO $\text{(VOC)}\rightarrow\text{COCO}$ (non-VOC) setting. *: Uses ground truth of unknown classes.}
\label{tab:study_quan_res}
\begin{tabular}{@{}lcccc|clccc@{}}
\toprule
Object Prior & AP & AR$_{100}$ & F$_1$ & & & Object Prior & AP & AR$_{100}$ & F$_1$\\ \midrule
SOS + \textit{Grid} & 3.8 & 36.5 & 6.9 & & & SOS + \textit{DeepGaze} & 5.4 & 35.9 & 9.4 \\
SOS + \textit{Dist} & 3.4 & 27.4 & 6.0 & & & SOS + \textit{CAM} & 5.4 & 36.7 & 9.4 \\
SOS + \textit{Spx} & 5.6 & 34.8 & 9.6 & & & SOS + \textit{DINO} & \textbf{8.9 }& \textbf{38.1} & \textbf{14.4}\\
SOS + \textit{Contour} & 5.6 & 36.6 & 9.7 & & & SOS + \textit{U-Net} & 7.3 & 37.3 & 12.2\\
SOS + \textit{VOCUS2} & 6.1 & 37.7 & 10.5 & & & SOS + \textit{GT} & 18.1* & 42.5* & 25.4*\\
\bottomrule
\end{tabular}
\end{table}

\section{Evaluation}
In our evaluation, we compare SOS to recent OWIS methods OLN~\cite{kim2022learning}, LDET~\cite{saito2022learning}, GGN~\cite{wang2022open}, SWORD~\cite{wu2023exploring}, SOIS~\cite{xue2022single}, OpenInst~\cite{wang2023openinst}, UDOS~\cite{kalluri2023open}, and against Mask R-CNN~\cite{he2017mask} and SAM~\cite{kirillov2023segment} baselines, depending on the availability of public source code or reported results. To assess the quality of the approaches, we use four cross-category and cross-dataset settings, commonly used in OWIS. Specifically, we choose COCO $\text{(VOC)}\rightarrow\text{COCO}$ (non-VOC) on the COCO dataset~\cite{lin2014microsoft}, already discussed in~\cref{sec:objPrior_eval_setup}. For cross-dataset evaluation, we use $\text{COCO}\rightarrow\text{LVIS}$,  $\text{COCO}\rightarrow\text{ADE20k}$, and $\text{COCO}\rightarrow\text{UVO}$ that train on the entire COCO training set and evaluate on the validation sets of LVIS~\cite{gupta2019lvis}, ADE20k~\cite{zhou2019semantic}, and UVO~\cite{wang2021unidentified} with exhaustive annotations. The LVIS dataset covers the same images as COCO but extends the annotated object classes from 80 to 1203. ADE20k includes 3169 classes of objects, object parts, and stuff regions. Finally, UVO features exhaustive annotations of all objects and, therefore, no system is penalized for detecting objects outside a datasets' taxonomy.

As in~\cref{sec:objPrior_eval_setup} and following common practice in OWIS~\cite{saito2022learning,wang2022open,kim2022learning}, we report AR for up to 100 detections and AP. We again also report the F$_1$ score between AR and AP as a single, combined quantity, described in~\cref{sec:objPrior_eval_setup}. Note that we only evaluate based on masks, not boxes.


\subsection{Implementation Details and Data Usage}
As discussed in~\cref{sec:objPrior_eval}, SOS uses the object prior \textit{DINO} for best results. DINO~\cite{caron2021emerging} with a ViT-S backbone is used as an ImageNet~\cite{russakovsky2015imagenet} pre-trained model without fine-tuning. Similarly, SOS uses SAM~\cite{kirillov2023segment} with the ViT-H backbone as a pre-trained model based on SA-1B~\cite{kirillov2023segment} without fine-tuning. As an instance segmentation model, we apply  Mask R-CNN~\cite{he2017mask} with an ImageNet pre-trained ResNet-50+FPN backbone~\cite{lin2017feature}, and the default configurations except for the class-agnostic training. 

Overall, SOS is based on models pre-trained on standard ImageNet and SA-1B. Pre-training on SA-1B or similar large-scale datasets becomes common with the emergence of foundation models and is widely adopted by systems using SAM in similar contexts~\cite{chen2023segment,yang2024foundation,wei2023semantic}. It is comparable to using models pre-trained on ImageNet that even have class information available during training and also allow a rough localization of objects~\cite{zhou2016learning,caron2021emerging}. Moreover, the final, noisy annotations of SA-1B representing objects, object parts, and stuff region were automatically generated by SAM~\cite{kirillov2023segment}, and no data leakage w.r.t. other datasets exists. 


\subsection{Cross-category COCO $\text{(VOC)}\rightarrow\text{COCO}$ (non-VOC) Results}
Table~\ref{tab:eval_coco_voc} presents the results of various OWIS methods on the cross-category setup COCO $\text{(VOC)}\rightarrow\text{COCO}$ (non-VOC). Our SOS clearly outperforms all OWIS methods in AR$_{100}$, AP, and F$_1$. Specifically, SOS outperforms the second-best system in terms of F$_1$, GGN~\cite{wang2022open}, by 6.1. While some of the improvement comes from a better recall~(+38.9\% compared to GGN), the results are driven by a much-improved precision~(+81.6\% compared to GGN). This reflects the high quality of the pseudo annotations in SOS on this cross-category setting. Moreover, there is a clear improvement of SOS over Mask R-CNN~(+12.7 in F$_1$). Compared to original SAM, the \textit{DINO}-based SOS substantially improves the precision by focusing on objects, leading to an improved F$_1$ score~(+7.8). 

\begin{figure}[t]
    \centering
    \begin{tabular}{cccccc}
\includegraphics[width=.16\linewidth]{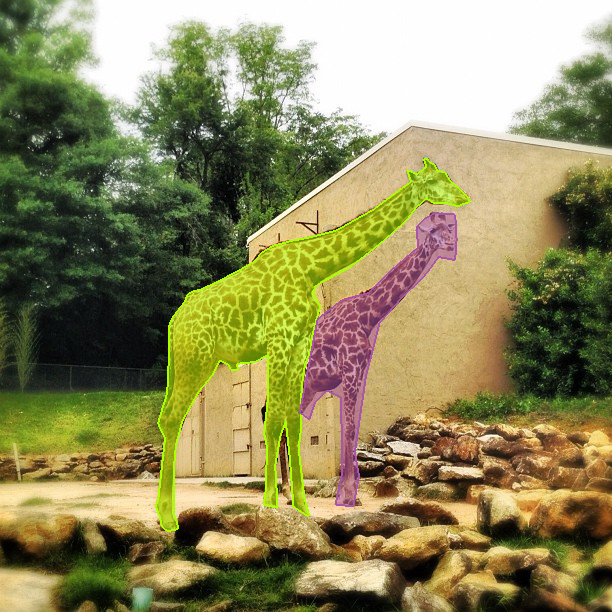} & \includegraphics[width=.16\linewidth]{figs/33707_baseline.png}  & \includegraphics[width=.16\linewidth]{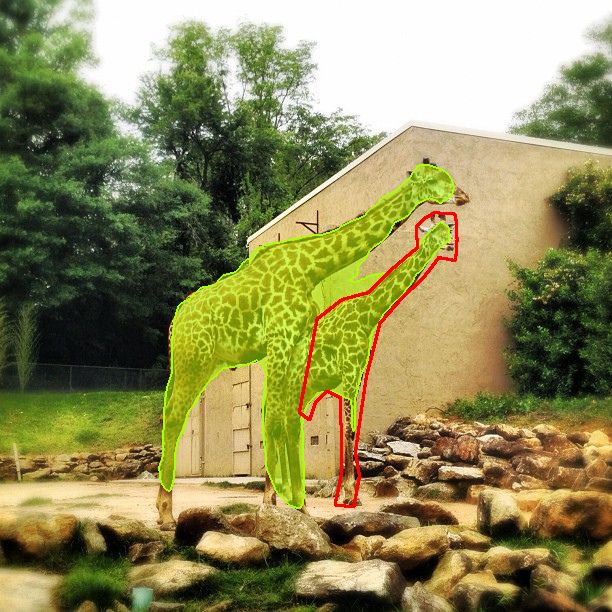} & \includegraphics[width=.16\linewidth]{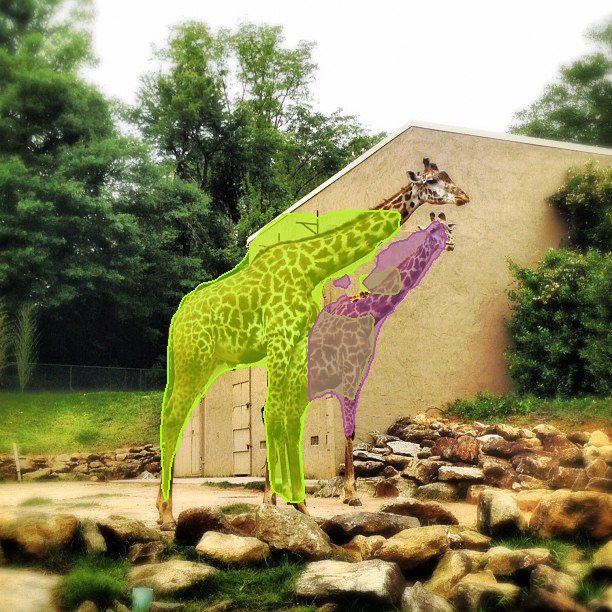} & \includegraphics[width=.16\linewidth]{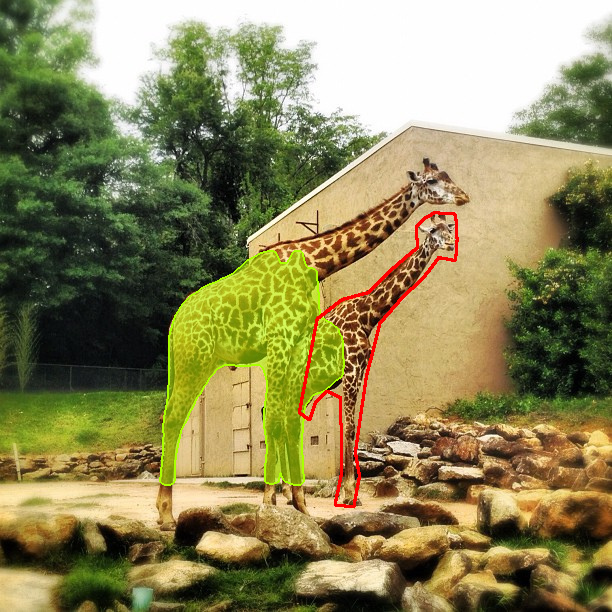} & \includegraphics[width=.16\linewidth]{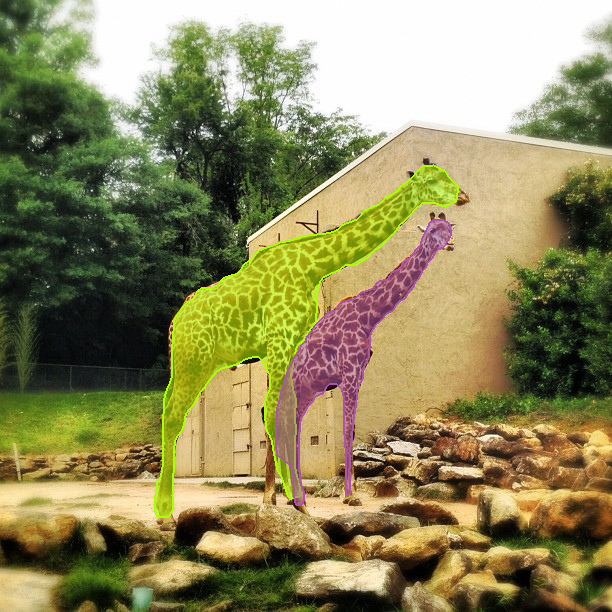} \\
 \includegraphics[width=.16\linewidth]{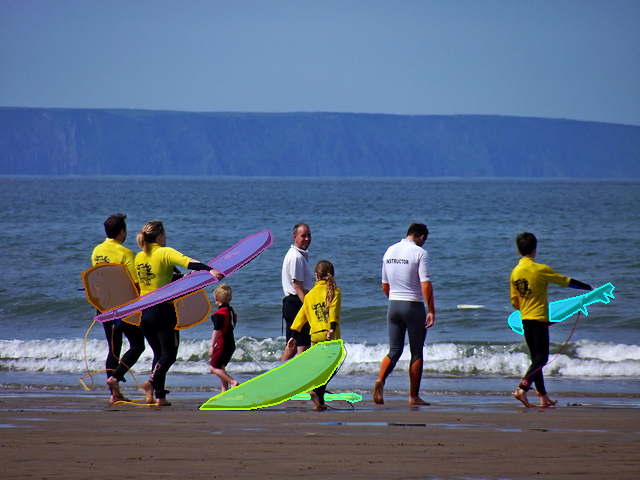} & \includegraphics[width=.16\linewidth]{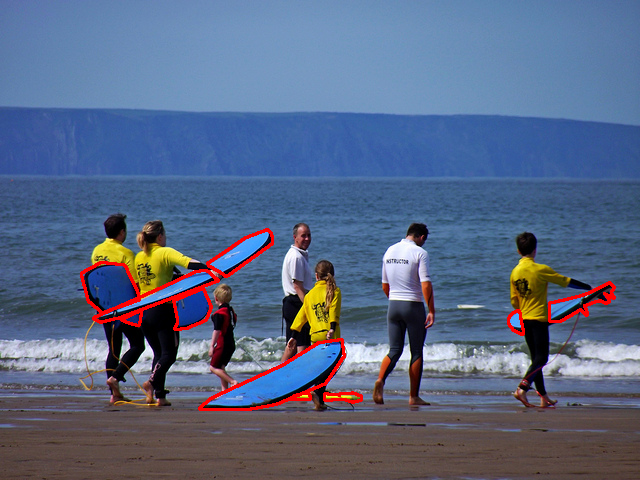} & \includegraphics[width=.16\linewidth]{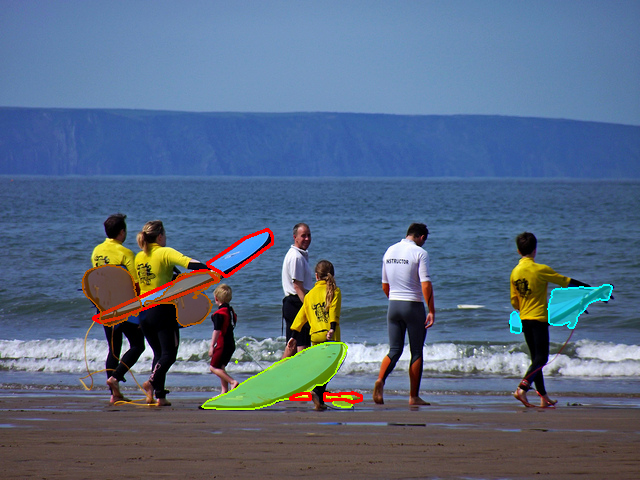} & \includegraphics[width=.16\linewidth]{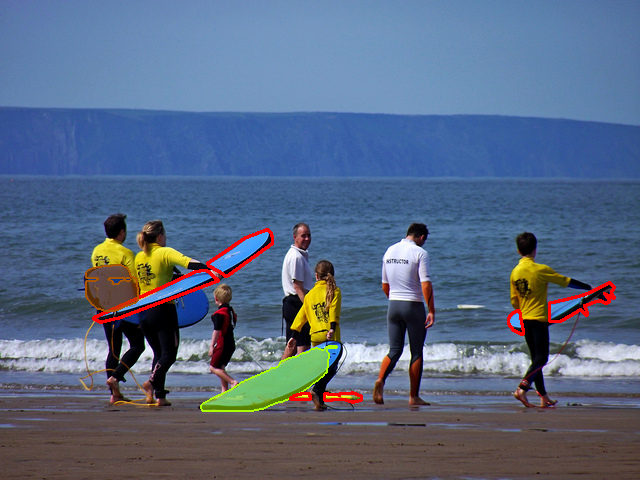} & \includegraphics[width=.16\linewidth]{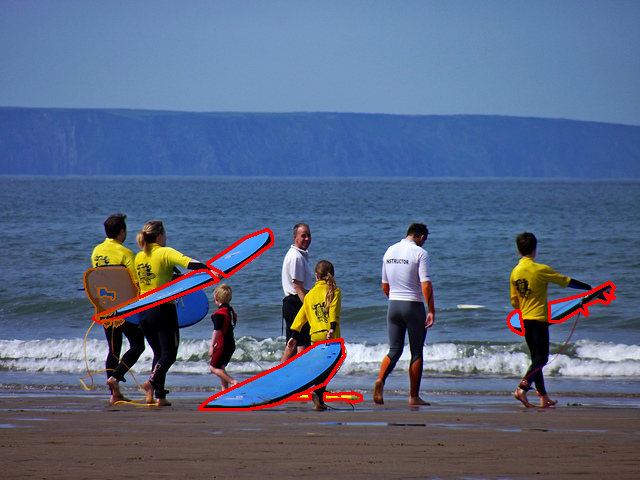} & \includegraphics[width=.16\linewidth]{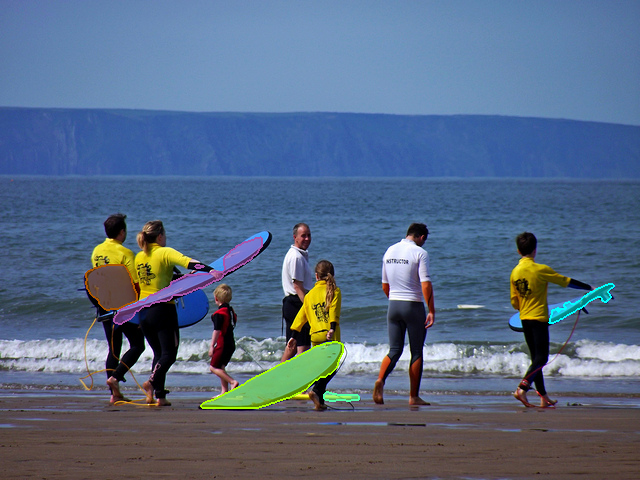}  \\
{\scriptsize Ground Truth} & {\scriptsize Mask R-CNN}  & {\scriptsize LDET} & {\scriptsize GGN} & {\scriptsize UDOS} & {\scriptsize SOS (ours)}
\end{tabular}
    \caption{Qualitative results of OWIS methods and baseline Mask R-CNN in the cross-category COCO $\text{(VOC)}\rightarrow \text{COCO}$ (non-VOC) setting. Filled masks denote detected objects, while red frames indicate missed objects.}
    \label{fig:qualRes}
\end{figure}

The qualitative results of various OWIS methods in~\cref{fig:qualRes} support the quantitative results. In the first example, only GGN~\cite{wang2022open} and SOS detect both giraffes, however, SOS generates more accurate segmentations. In the second example, only SOS detects all five small surfboards, while other systems detect at most three. Note that class-agnostic Mask R-CNN misses all non-VOC objects in both examples. For more qualitative results, see our supplementary.


\begin{table}[t]
\parbox{.48\linewidth}{
\centering
\caption{Results of two baselines and various OWIS methods in the COCO $\text{(VOC)}\rightarrow\text{COCO}$ (non-VOC) setting. $^\dagger$: uses automatically annotated SA-1B dataset.}
\label{tab:eval_coco_voc}
\begin{tabular}{@{}lccc@{}}
\toprule
System & AP & AR$_{100}$ & F$_1$ \\ \midrule
Mask R-CNN~\cite{he2017mask}& 1.0 & 8.2 & 1.8 \\ 
SAM$^\dagger$~\cite{kirillov2023segment} & 3.6 & \textbf{48.1} & 6.7 \\
\midrule
OLN~\cite{kim2022learning} & 4.2 & 28.4 & 7.3 \\
LDET~\cite{saito2022learning} & 4.3 & 24.8 & 7.3 \\
GGN~\cite{wang2022open} & 4.9 & 28.3 & 8.4 \\
SWORD~\cite{wu2023exploring} & 4.8 & 30.2 & 8.3 \\ 
UDOS~\cite{kalluri2023open} & 2.9 & 34.3 & 5.3 \\ \midrule
SOS$^\dagger$ (ours) & \textbf{8.9} & 39.3 & \textbf{14.5} \\ \bottomrule
\end{tabular}
}
\hfill
\parbox{.48\linewidth}{
\centering
\caption{Results of two baselines and various OWIS methods in the $\text{COCO} \rightarrow \text{LVIS}$ setting. $^\dagger$: uses automatically annotated SA-1B dataset.}
\label{tab:eval_lvis_coco}
\begin{tabular}{@{}lccc@{}}
\toprule
System & AP & AR$_{100}$ & F$_1$ \\ \midrule
Mask R-CNN~\cite{he2017mask}& 7.5 & 23.6 & 11.4 \\ 
SAM$^\dagger$~\cite{kirillov2023segment} & 6.8 & \textbf{45.1} & 11.8 \\ \midrule
LDET~\cite{saito2022learning} & 6.7 & 24.8 & 10.5 \\
GGN~\cite{wang2022open} & 6.5 & 27.0 & 10.5 \\
SOIS~\cite{xue2022single} & - & 25.2 & - \\ 
OpenInst~\cite{wang2023openinst} & - & 29.3 & - \\ 
UDOS~\cite{kalluri2023open} & 3.9 & 24.9 & 6.7 \\ \midrule
SOS$^\dagger$ (ours) & \textbf{8.1} & 33.3 & \textbf{13.3} \\ \bottomrule\\
\end{tabular}
}
\end{table}

\subsection{Cross-dataset Results}
\paragraph{\textbf{$\text{COCO}\rightarrow\text{LVIS}$:}}
We evaluate the cross-dataset generalization capabilities of OWIS methods, starting with $\text{COCO}\rightarrow\text{LVIS}$. The results in~\cref{tab:eval_lvis_coco} show a trend similar to the cross-category results. SOS outperforms all OWIS methods across all measures with an improvement of 2.8 in F$_1$ compared to second-best  methods GGN~\cite{wang2022open} and LDET~\cite{saito2022learning}. Similar to the previous results, the improvement is based on gains in both recall and precision, however, the relative improvements in AP and AR$_{100}$ are more similar to each other. Overall, our high-quality pseudo annotations in SOS lead to new state-of-the-art results, which indicate  strong generalization to unknown object classes outside COCO.

\begin{table}[t]
\parbox{.48\linewidth}{

\centering
\caption{Results of Mask R-CNN baseline and various OWIS methods in the $\text{COCO}\rightarrow\text{ADE20k}$ setting. $^\dagger$: uses automatically annotated SA-1B dataset.}
\label{tab:eval_ade20k}
\begin{tabular}{@{}lccc@{}}
\toprule
System & AP & AR$_{100}$ & F$_1$ \\ \midrule
Mask R-CNN~\cite{he2017mask}& 6.9 & 11.9 & 8.7 \\ \midrule
OLN~\cite{kim2022learning} & - & 20.4 & - \\ 
LDET~\cite{saito2022learning} & 9.5 & 18.5 & 12.6 \\
GGN~\cite{wang2022open} & 9.7  & 21.0 & 13.3 \\
UDOS~\cite{kalluri2023open} & 7.6 & 22.9 & 11.4 \\ \midrule
SOS$^\dagger$ (ours) & \textbf{12.5} & \textbf{26.5} & \textbf{17.0} \\ \bottomrule
\end{tabular}
}
 \hfill
\parbox{.48\linewidth}{
\centering
\caption{Results of two baselines and various OWIS methods in the $\text{COCO}\rightarrow\text{UVO}$ setting. $^\dagger$: uses automatically annotated SA-1B dataset.}
\label{tab:eval_uvo}
\begin{tabular}{@{}lccc@{}}
\toprule
System & AP & AR$_{100}$ & F$_1$ \\ \midrule
Mask R-CNN~\cite{he2017mask}& 20.7 & 36.7 & 26.5 \\ 
SAM$^\dagger$~\cite{kirillov2023segment} & 11.3 & \textbf{50.1} & 18.4 \\ \midrule
OLN~\cite{kim2022learning} & - & 41.4 & - \\ 
LDET~\cite{saito2022learning} & \textbf{22.0} & 40.4 & \textbf{28.5 }\\
GGN~\cite{wang2022open} & 20.3 & 43.4 & 27.7 \\
UDOS~\cite{kalluri2023open} & 10.6 & 43.1 & 17.0 \\ \midrule
SOS$^\dagger$ (ours) & 20.9 & 42.3 & 28.0 \\ \bottomrule
\end{tabular}

}
\end{table}

\paragraph{\textbf{$\text{COCO}\rightarrow\text{ADE20k}$:}}
The results for $\text{COCO}\rightarrow\text{ADE20k}$ in~\cref{tab:eval_ade20k}, which include labeled object parts of ADE20k, show that SOS again outperforms all other OWIS methods across all measures. The gain over the second-best method in F$_1$, GGN~\cite{wang2022open}, is 3.7. Similar to $\text{COCO}\rightarrow\text{LVIS}$, the relative gains in AR$_{100}$ and AP are equally distributed. These results indicate that object parts do not degrade the results of SOS. This is in line with SOS's pseudo annotations in~\cref{fig:qual_priors}, covering entire objects and selected object parts.

\paragraph{\textbf{$\text{COCO}\rightarrow\text{UVO}$:}}
Finally, \cref{tab:eval_uvo} presents results for $\text{COCO}\rightarrow\text{UVO}$. SOS still outperforms Mask R-CNN, SAM and recent UDOS~\cite{kalluri2023open} in F$_1$. Outperforming SAM implies that SOS does not exploit the limited taxonomy of annotations in COCO (only 60 object classes in test), but also improves the segmentation of objects outside the COCO object classes on the exhaustively labeled UVO dataset. However, the results of SOS are below LDET~\cite{saito2022learning} in AP and F$_1$. We attribute this to several classes in UVO not available in ImageNet, which was used to learn \textit{DINO} object prior. Hence, \textit{DINO} will miss such objects in the COCO dataset during training, and SOS might miss them during testing. This could be mitigated using more diverse unlabeled images for training \textit{DINO}.

\subsection{Quality of Pseudo Annotations}

We also investigate the quality of the pseudo annotations generated in SOS. To this end, we evaluate the pseudo annotations on the COCO training set against the annotations of non-VOC object classes, reflecting the COCO $\text{(VOC)}\rightarrow\text{COCO}$ non-VOC) setting. The results in terms of precision and recall for $\text{IoU} = 0.5$ as well as the F$_1$ score are presented in~\cref{tab:eval_qualAnns}. We use up to three and 10 pseudo annotations generated per image in SOS~(SOS$_3$ and SOS$_{10}$), and GGN$_3$~\cite{wang2022open} that generates three pseudo annotations. The results show that with only three SOS pseudo annotations, more than 25\% of the non-VOC objects are covered. This is substantially more than GGN$_3$~(12.1\%). With up to 10 annotations, SOS's pseudo annotations cover more than 40\% of the non-VOC objects. In our supplementary, we further provide class-specific results.

To complement these results, we visualize the annotations of GGN$_3$ and SOS$_3$ on two sample images in~\cref{fig:gtExamples}. It is visible that GGN$_3$'s annotations cover background regions, leading to low precision in the overall results. Moreover, the pseudo annotations in the upper example do not adhere well to the airplane's boundaries. More qualitative examples are given in~\cref{fig:intro} and our supplementary.

\begin{table}[t]
\parbox{.48\linewidth}{
\vspace*{-2em}
\centering
\caption{Quality of pseudo annotation generated in GGN~\cite{wang2022open} and our SOS evaluated on the non-VOC annotations of the COCO training dataset. $^\dagger$: uses automatically annotated SA-1B dataset.}
\label{tab:eval_qualAnns}
\begin{tabular}{@{}lccc@{}}
\toprule
System & Prec. & Rec. & F$_1$ \\ \midrule
GGN$_3$~\cite{wang2022open} & 7.3 & 12.1 & 9.1 \\ \midrule
SOS$_3^\dagger$ (ours) & \textbf{19.0} & 26.4 & 22.1 \\
SOS$_{10}^\dagger$ (ours) & 15.5 & \textbf{41.7} & \textbf{22.6} \\ \bottomrule 
\end{tabular}
}
\hfill
\parbox{.48\linewidth}{
\centering
    \begin{tabular}{cc}
\includegraphics[width=0.4\linewidth]{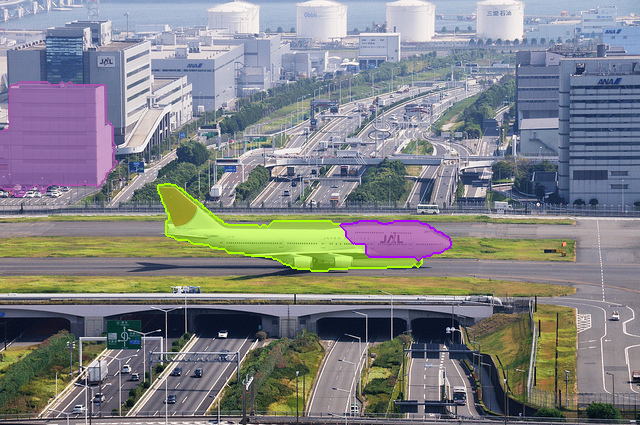} & \includegraphics[width=0.4\linewidth]{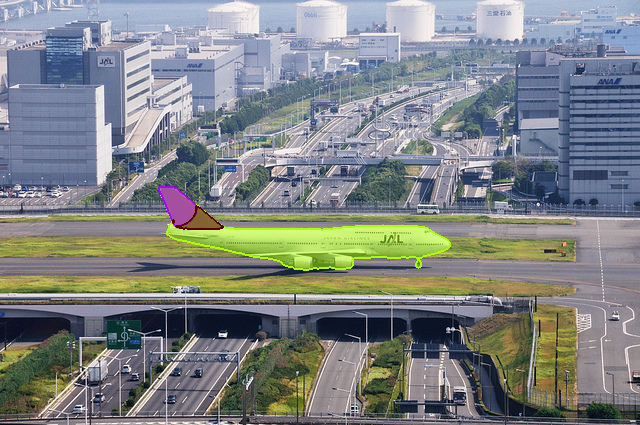} \\
\includegraphics[width=0.4\linewidth]{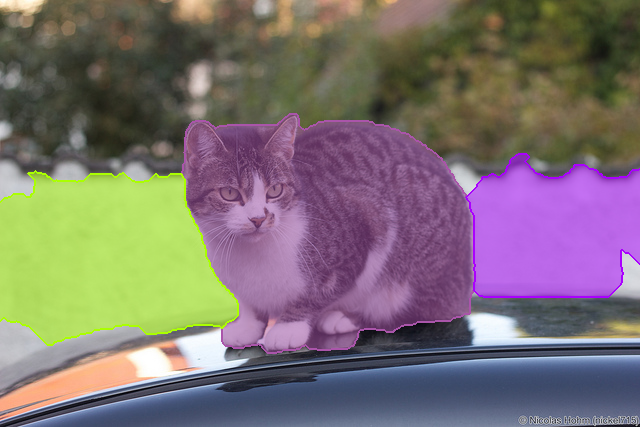} & \includegraphics[width=0.4\linewidth]{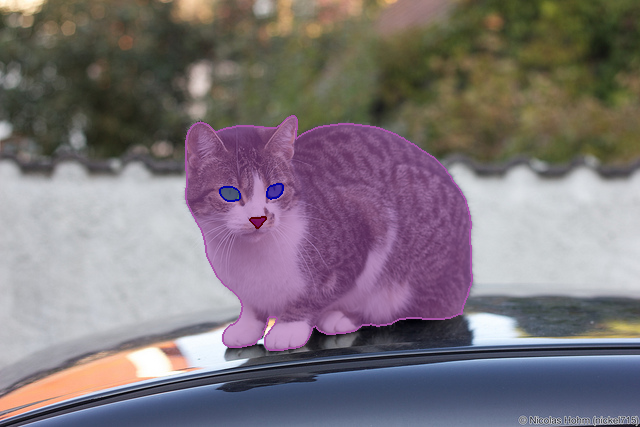} \\
{\scriptsize GGN} & {\scriptsize SOS}
\end{tabular}
    \captionof{figure}{Pseudo annotations generated by GGN and our SOS.}
    \label{fig:gtExamples}
}
\end{table}



\subsection{Ablation Studies}
All ablation studies follow the evaluation setup described in~\cref{sec:objPrior_eval_setup}. More studies evaluating  design choices in SOS are presented in our supplementary.


\paragraph{\textbf{Influence of SOS's Components:}}
We investigate the importance of each component in SOS and present the results in~\cref{tab:eval_abl}. The baseline for this study is a class-agnostic Mask R-CNN trained without pseudo annotations~(first row in~\cref{tab:eval_abl}). Subsequently, we add pseudo annotations based on SAM with the baseline \textit{Grid} object prior as in~\cite{kirillov2023segment} and without our post-processing in PAC~(confidence-based thresholding and NMS, see~\cref{sec:method_pac}). Next, we replace \textit{Grid} with \textit{DINO}~(third row in~\cref{tab:eval_abl}), and finally add the post-processing leading to the complete SOS. The results in~\cref{tab:eval_abl} show each step improves the F$_1$ results. Mainly, introducing SAM for generating pseudo annotations (first row vs. second row) substantially improves AR$_{100}$, while adding the \textit{DINO} object prior removes background annotations leading to considerably improved precision. Hence, the object prior selection in SOS is crucial for strong results.


\begin{table}[t]
\parbox{.48\linewidth}{
\centering
\caption{Influence of added components in our SOS evaluated in the COCO $\text{(VOC)}\rightarrow\text{COCO}$ (non-VOC) setting.}
\label{tab:eval_abl}
\begin{tabular}{@{}lccc@{}}
\toprule
Components & AP & AR$_{100}$ & F$_1$ \\ \midrule
Mask R-CNN & 1.2 & 10.8 & 2.2 \\ 
+ \textit{Grid}, w/o pp. & 3.4 & 35.2 & 6.2 \\
+ \textit{DINO}, w/o pp. & \textbf{8.9} & 37.1 & 14.3 \\
+ \textit{DINO}, w/ pp. & \textbf{8.9} & \textbf{38.1} & \textbf{14.4} \\\bottomrule
\end{tabular}
}
\hfill
\parbox{.48\linewidth}{
\centering
\caption{SOS results with various numbers of pseudo annotations in the COCO $\text{(VOC)}\rightarrow\text{COCO}$ (non-VOC) setting.}
\label{tab:eval_numAnns}
\begin{tabular}{@{}lccc@{}}
\toprule
\#Pseudo Anns. & AP & AR$_{100}$ & F$_1$ \\ \midrule
3 & 8.3 & 34.5 & 13.4 \\ 
5 & 8.8 & 36.2 & 14.2 \\
10 & \textbf{8.9} & \textbf{38.1} & \textbf{14.4} \\
20 & 8.8 & \textbf{38.1} & 14.3 \\\bottomrule
\end{tabular}
}
\end{table}


\paragraph{\textbf{Number of Pseudo Annotations:}}
In the second ablation study, we investigate the influence of the number of pseudo annotations per image on the results of SOS. To this end, we evaluate SOS with up to 3, 5, 10, and 20 pseudo annotations per image. The results in~\cref{tab:eval_numAnns} show that 10 pseudo annotations are preferable. Hence, the ideal number of pseudo annotations in SOS is larger than in GGN~\cite{wang2022open}, which only uses three. This indicates again the high quality of our pseudo annotations, since more than three annotations per image can cover relevant objects without adding too many noisy annotations.




\section{Conclusion}
In this paper, we addressed the challenging OWIS task and aimed to improve the low precision of previous methods in this open-world task. To improve the precision and the overall detection results, we generate high-quality pseudo annotations based on a prompt-guided foundation model, SAM. To focus the pseudo annotations on objects, we thoroughly investigated various object priors for prompt creation, which revealed important insights on how to prompt SAM for class-agnostic object-focused results in arbitrary tasks. As a result, our novel OWIS method SOS, which uses a self-attention-based object prior for generating pseudo annotations, outperforms recent state-of-the-art OWIS methods across challenging open-world setups in COCO, LVIS, and ADE20k datasets. SOS especially improves the precision based on the high-quality pseudo annotations. Moreover, SOS with its focus on objects also outperforms vanilla SAM that segments both object and stuff regions.


%
%
\bibliographystyle{splncs04}
\bibliography{lit}
\end{document}